\documentclass{article}

\usepackage[comma]{natbib}
\usepackage{url}
\usepackage{graphicx}

\usepackage{xcolor}
\usepackage{paralist}
\usepackage{amsfonts,amssymb,amsmath}
\usepackage{natbib}
\usepackage{url}
\usepackage{booktabs}

\usepackage{pifont}%
\newcommand{\cmark}{\ding{51}}%

\renewcommand{\le}{\leqslant}

\renewcommand{\emptyset}{{\varnothing}}

\newcommand{\giv}{\!\mid\!}

\newcommand{\newt}[1]{#1} %

\newcommand{\wh}{\widehat}
\newcommand{\real}{\mathbb{R}}

\newcommand{\bsx}{\boldsymbol{x}}

\newcommand{\bsz}{\boldsymbol{z}}

\newcommand{\val}{\nu}

\newcommand{\e}{\mathbb{E}}
\newcommand{\var}{\mathrm{var}}
\newcommand{\rd}{\,\mathrm{d}}

\newcommand{\olt}{\overline\tau}
\newcommand{\ult}{\underline\tau}

\newcommand{\phm}{\phantom{-}}

\newcommand{\cx}{\mathcal{X}}

\newcommand{\fp}{\mathrm{FP}}
\newcommand{\fn}{\mathrm{FN}}
\newcommand{\sumdot}{\text{\tiny$\bullet$}}

\newcommand{\fpr}{\mathrm{FPR}}
\newcommand{\fnr}{\mathrm{FNR}}

\newcommand{\ppv}{\mathrm{PPV}}

\newcommand{\dunif}{\mathsf{U}}
\usepackage{geometry}
\usepackage{url}

\begin{document}

\title{Variable importance without impossible data}

\author{
   Masayoshi Mase\\Hitachi, Ltd.
   \and
   Art B. Owen \\ Stanford University
   \and
   Benjamin B. Seiler \\ Stanford University
}
\date{April 2023}
\maketitle

\begin{abstract}
The most popular methods for measuring importance
of the variables in a black-box prediction algorithm
make use of synthetic inputs that combine predictor variables from multiple observations.
These inputs can be unlikely, physically impossible, or even logically impossible.
As a result, the predictions for such cases can be based on data very unlike any the black box was trained on.
We think that users cannot trust an explanation of the decision of a prediction algorithm when the explanation uses such values.
Instead we advocate a method called Cohort Shapley that is grounded 
in economic game theory and uses only actually observed data to quantify variable importance.
Cohort Shapley works by narrowing the cohort of observations judged to be similar to a target observation on one or more features.
We illustrate it on an algorithmic fairness problem where it is essential to attribute importance to protected variables that the model was not trained on.
\end{abstract}

\section{Introduction}

Black-box models commonly attain state of the art prediction accuracy as measured by predictions of held-out response values. In many settings, it is not enough to know that they are accurate. We want to interpret them too. We might derive a useful scientific insight from understanding how a prediction works, or we might spot a flaw in the model that convinces us that it will not generalize well beyond our data set. We might also find that the algorithm works in a way that is unfair to some people.

A first step in interpreting a black-box model is to discern what are the important variables in that model. This is already a hard problem, and as we note below there are several large branches of the literature devoted to this problem. Most of the methods in those areas work by making changes to some, but not necessarily all, of the input variables in a model and taking note of how the predictions change in response. A severe problem with those approaches is that they will use unlikely or even impossible combinations of inputs. For instance, if one combines the youngest age of a criminal defendant with the greatest observed number of prior arrests, the result is an extremely unlikely synthetic data point. Combining features from different observations can lead to synthetic points describing people who graduated from high school before they were born, or who live in both Idaho and Los Angeles County. In medical settings, one can get impossible patients with systolic blood pressures below their diastolic blood pressure, or whose minimum O$_2$ saturation level is above their average. 

\begin{figure}
    \centering
    \includegraphics[width=.9\hsize]{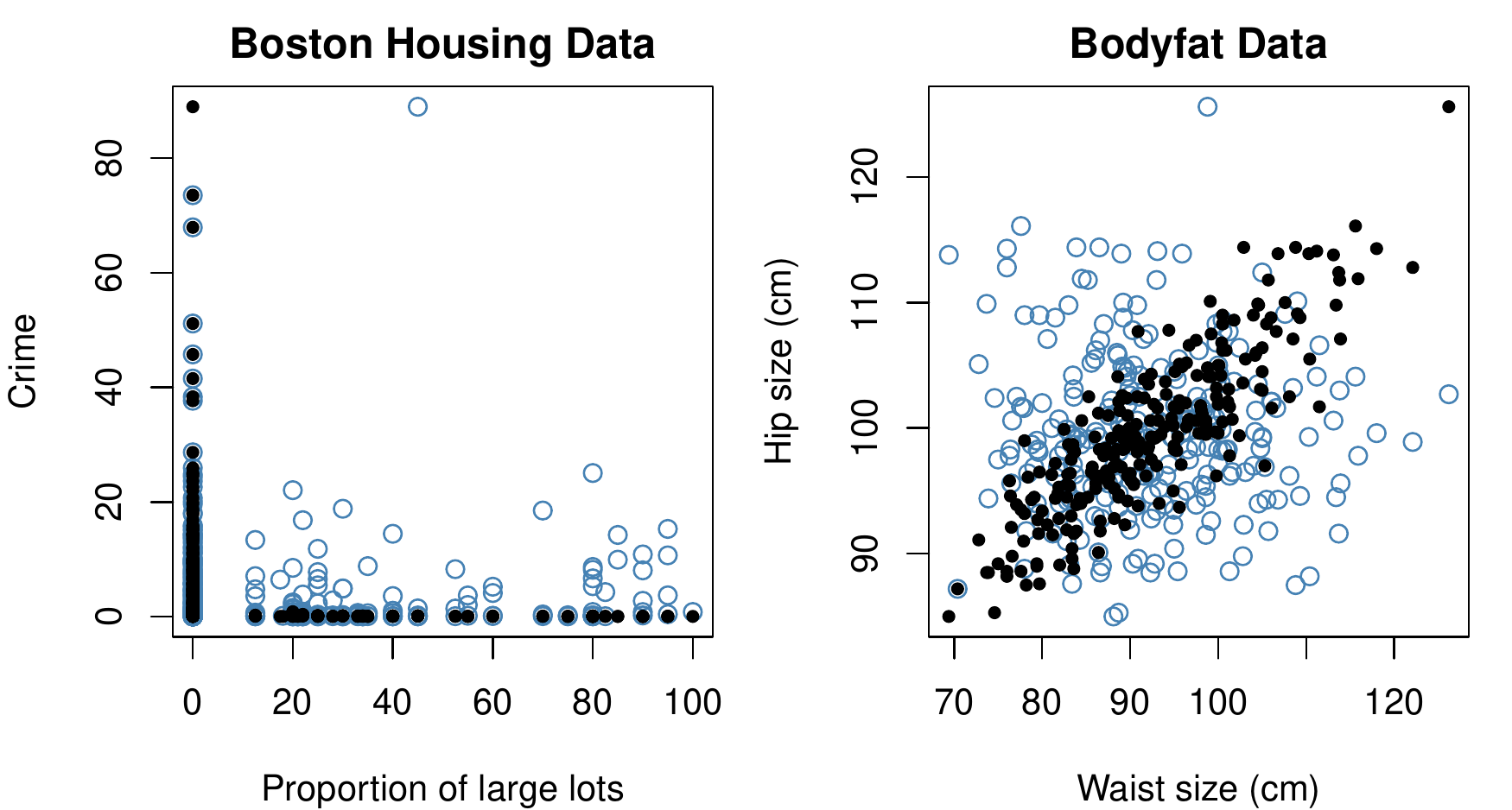}
    \caption{
    \label{fig:dontpermute} The left panel plots the crime variable versus the proportion of large lots for the Boston housing data. The right panel plots hip size versus waist size in the bodyfat data. In both panels, solid points are real observations and open circles have one variable permuted with respect to the other.
    }
\end{figure}

Figure~\ref{fig:dontpermute} shows what this can look like when levels of one variable are permuted with respect to levels of another variable. The left panel uses the well known Boston housing data of \citet{harr:rubi:1978}. The right panel uses the bodyfat data from the Data Analysis and Story Library (\url{https://dasl.datadescription.com}) after converting the waist measurements there from inches to centimeters. 
We see that the synthetic variable combinations can be quite unlike the real data. Those combinations may simply have nothing to do with any parts of Boston or with any likely human physique. This poses a problem: a black-box prediction function evaluated at such points is then being used far from where it was trained. Such extrapolations are at a minimum unreliable, and potentially meaningless. Furthermore, there are adversarial attacks on variable importance measures that use such inputs. \citet{slac:etal:2020} report how an algorithm that makes unfair predictions at observed data values can be rated as fair due to predictions made at synthetic data points formed by mixing and matching inputs. This applies to two of the leading methods in machine learning: the Local Interpretable Model agnostic Interpretation (LIME) of \citet{ribe:etal:2016} and the SHapley Additive exPlanations (SHAP) of \citet{lund:lee:2017}. In contrast, the TreeSHAP method \citep{lund:2020}
uses only observed values; we describe below how it differs from our proposal. For now, we note that it is only defined for tree structured predictions.

For the reasons above, we need a variable importance measure that does not use impossible data. While there are algorithmic methods to judge when a data point is `out of distribution,' there is not a reliable way to say exactly where the support of a distribution ends. In this article, we present the cohort Shapley measure from \citet{mase:owen:seil:2019} as one solution. Like many of the methods we discuss, this algorithm is grounded in axioms of economic game theory from \citet{shap:1953}. However, the way the underlying game value is defined allows us to only use observed combinations of predictor variables.

This article is organized as follows: 
\newt{Section~\ref{sec:compandexample} compares 
properties of cohort Shapley to
some widely adopted variable importance methods and
introduces a running example for illustration.}
Section~\ref{sec:varimpgeneral} provides a brief discussion of the problems of variable importance and points to literature surveys in three areas: statistics, machine learning and global sensitivity analysis. At first sight, variable importance looks to be easy because all the counterfactuals we might want to consider can usually be directly computed. At a second look, variable importance is actually extremely difficult because it seeks to identify causes of effects instead of effects of causes. \newt{We explain that distinction making reference to \cite{dawi:musi:2021}.} Section \ref{sec:techniques} presents functional ANOVA, Sobol' indices from global sensitivity analysis, and Shapley value from game theory. Section \ref{sec:shapley4importance} describes how Shapley value has been used to quantify variable importance and presents the cohort Shapley method. Section \ref{sec:compas} illustrates cohort Shapley on an algorithmic fairness problem using data from \citet{angw:2016} on the COMPAS algorithm for predicting recidivism. The key issue there is whether Black defendants were treated unfairly. For every defendant, cohort Shapley can estimate the importance to their prediction of each of their feature variables. We are able to do this even though the algorithm was trained without using race as a variable. Furthermore, we are able to do this just using the algorithm's predictions, which is an advantage because the algorithm is proprietary and not available to us. We can aggregate importance over subsets of observations, consistent with an additivity axiom. We also show how to use the Bayesian bootstrap of \citet{rubi:1981} to get confidence intervals for variable importance of individuals and of aggregates. 
Section~\ref{sec:discussion} has some final remarks.
Python code for the cohort Shapley method is available
at  \url{https://github.com/cohortshapley/cohortshapley}.

\section{Comparisons and a running example}\label{sec:compandexample}
Here we compare cohort Shapley
to some other methods that we will describe
in later sections.
Variable importance methods differ primarily in
how they define importance. We base our comparisons
on which useful properties they have or do not have,
as shown in Table~\ref{tab:affordances}.
We have phrased the properties in a way
that makes positive values desirable.   Attributing
importance to variables that are not in the model
is controversial.  We think that
the ability to do that is desirable; cohort Shapley allows that but does not require it.  TreeSHAP is fully automatic.
Baseline Shapley and most other methods require
defining a baseline point to compare the target point
to.  Cohort Shapley also requires a user-provided
notion of variable similarity. We have listed LIME as
requiring a second black box because it has a regularization
process in determining how local an observation must
be to be similar to the target.
Kernel SHAP is listed as not computing the Shapley
value because it computes a kernelized approximation
to it.  Similarly, on-manifold SHAP and 
conditional kernel SHAP do not compute the Shapley value; cohort
and baseline Shapley do compute it.  We include Monte
Carlo versions of them because they are consistent
for the Shapley value as computation increases.
LIME requires the choice a surrogate
model and a kernel, so we do not consider it to be automatic.

\begin{table}[t]
    \centering
    \begin{tabular}{lcccccccccc}
    \toprule
    Property & Cohort & Baseline & Tree & Kernel & IG & Manifold & Cond & LIME\\
    \midrule
    Real obs. only& \cmark & &\cmark& & &&&\\
    Allows $x$ dependence&\cmark&&&&&\cmark&\cmark&\\
    Unused variables &\cmark& &&&&\cmark&&\\
    Does not need $f(\cdot)$ &\cmark&&&&&&&\\
    Uses any model &\cmark &\cmark&&\cmark&&\cmark&\cmark&\cmark\\
    Sub-exponential cost &&&\cmark&\cmark&\cmark&\cmark&\cmark&\cmark\\
    Computes Shapley &\cmark&\cmark&\cmark&&&&&\\
    No 2nd black box &\cmark&\cmark&\cmark&&\cmark&&&\\
    Automatic && &\cmark&&&&&&\\
    \bottomrule
    \end{tabular}
    \caption{
    \label{tab:affordances}
    Summary of properties of variable importance methods: cohort Shapley, baseline Shapley, TreeSHAP,  KernelSHAP,  integrated gradients, On-manifold SHAP and Conditional Kernel SHAP  described in Section \ref{sec:shapley4importance}
    and LIME of  \citet{ribe:etal:2016}.
    Properties from top to bottom:
    uses only observed predictor combinations,
    uses no independence assumption on inputs,
    can attribute importance to unused variables,
    does not need $f(\cdot)$ just its predictions,
    works with any model, cost of exact
    computation is not
    exponential in $d$,
    computes a Shapley value instead of
    an approximation,
    does not use a second black box to explain $f$, and
    requires no user input.
    }
\end{table}

Of the methods in Table~\ref{tab:affordances}, only
cohort Shapley can be used on the COMPAS data that consist only of the predictions from a proprietary algorithm with no information on the model itself. It is similarly the only method that works with residuals to compare actual versus predicted
levels of recidivism.
\newt{The residuals are important because they let us study
how an algorithm unfairly over-estimates or under-estimates
a risk. Also if a variable is important for the residual it
suggests that the model did not use that variable well.}
Cohort Shapley is one of only two that can attribute
importance to an unused variable and one of
only two that never uses unobserved data.

All the methods that we compare in Table~\ref{tab:affordances} can be used without retraining the model on different
feature sets.
They are aimed
at attributing a specific point's prediction to
one of their features, which is the task we consider
in our numerical examples.  Retraining with and without
certain features present has been used, for example
by \cite{strumbelwithretrainingj2009explaining}
and by \cite{hooker2019benchmark}
for their RemOve And Retrain (ROAR) method.

\newt{Table~\ref{tab:runningexample} shows a small example that we will use
to illustrate cohort Shapley.  It has three binary predictors, two
data distributions (one dependent and one independent) and two
response functions (one additive and one with interaction).
The dependent data distribution features an impossible combination.
We can think of a setting where one binary variable specifies
whether somebody has any children and a second whether their
first child is male. Then the $(0,1)$ combination is impossible.
Or we can think of a setting for the Boston housing data
describing a region with many large lots and also high crime
which while not logically impossible is quite improbable.
In the given example, we have made a three variable combination
impossible. We have also left variable three out of $f_1$.
Variable 3 could be something like the race of a person which
is not used by the algorithm at all, but might have
some importance from association
with variables that are used.}

\begin{table}\centering
\begin{tabular}{ccccccccc}
\toprule
Obs & $x_1$ & $x_2$ & $x_3$ & $P$ & $P_*$ &$f_1(\bsx)=x_1-2x_2$& $f_2(\bsx)=x_1x_2(1-x_3)$\\
\midrule
$t=1$ &0 & 0 & 0 & 1 & 1 & $\phm0$ &0\\
   2 &0 & 0 & 1 & 1 & 1 & $\phm0$ &0\\
   3 &0 & 1 & 0 & 1 & 1 & $-2$&0\\
   4 &0 & 1 & 1 & 1 & 1 & $-2$&0\\
   5 &1 & 0 & 0 & 1 & 1 & $\phm1$&0\\
   6 &1 & 0 & 1 & 1 & 1 & $\phm1$&0\\
   7 &1 & 1 & 0 & 1 & 0 & $-1$&1\\
   8 &1 & 1 & 1 & 1 & 1 & $-1$&0\\
\bottomrule
\end{tabular}
\caption{\label{tab:runningexample}
Here is a running example with $3$
binary variables.  The observation $t=1$
is the target for local explanation. 
In one setting denoted by $P$, all 8 possible
input values are equally common/probable,
so the variables are independent.
In another setting, denoted by $P_*$, the
variables are dependent because the $(1,1,0)$
combination is impossible. We consider 
two functions of $\bsx$: $f_1$ is linear and
does not involve $x_3$ while $f_2$ has interactions
and indicates the impossible $\bsx$.
}
\end{table}

\section{Variable importance}\label{sec:varimpgeneral}

We begin with some notation. We will consider a data set with $n$ observations indexed by $i\in1{:}n\equiv\{1,\dots,n\}$. For observation $i$ there is a $d$-tuple $\bsx_i=(x_{i1},\dots,x_{id})$ of predictor variables. The component variables $x_{ij}$ do not always have to be real-valued so $\bsx_i$ is not necessarily a vector in $\real^d$. The data set also includes a response $y_i$ for observation $i$. The models we consider predict $y_i$ from $\bsx_i$. We focus on the case $y_i\in\real$. \newt{Sometimes we omit the observation's index and speak of predicting
$y$ from $\bsx$. In some contexts we care about explaining
the predictions for one specific target who
then has index $t$.}

In our formal discussions, we will need some special notation to study the changes that methods make to $\bsx$. For $u\subseteq1{:}d$ we let $\bsx_u$ be the tuple of variables $x_j$ for $j\in u$. We will use $|u|$ for the cardinality of $u$ and $-u=1{:}d\setminus u$ for the complement of $u$. For tuples $\bsx$ and $\tilde\bsx$ we let $\bsx_u{:}\tilde\bsx_{-u}$ be the hybrid point $\bsx'$ with $\bsx'_j=\bsx_j$ for $j\in u$ and $\bsx'_j=\tilde\bsx_j$ for $j\not\in u$. We abbreviate $\bsx_{\{j\}}$ and $\bsx_{-\{j\}}$ to $\bsx_j$ and $\bsx_{-j}$, respectively.

\subsection{Goals and simple examples}
Our goal is to quantify the importance of the individual variables in a model that predicts $y$ from $\bsx$. Importance can be interpreted several ways. For instance, changing $x_{ij}$ could have a causal impact on $y_i$, removing predictor $x_j$ from a model to predict $y$ could degrade the accuracy of those predictions, or changing $x_{ij}$ could bring a large change to the prediction $\hat y_i$ \citep{jian:owen:2003}. These are distinct issues about how $x_j$ relates to $y$. We are most interested in understanding specific decisions made by a given algorithm. For instance, which variables best explain why an algorithm may have recommended against lending money to \newt{a target} applicant $t$, or may have recommended in favor of sending patient $t$ to an intensive care unit? For problems of this type, it is the third measure of importance that we want to study. In particular, an algorithm might have placed significant importance on a predictor known to have no causal effect, and we would want to detect that by showing how important that variable was to the predictions. It is for this goal that we especially want a method that avoids using impossible input combinations. In our conclusions we consider the consequences of including impossible combinations when judging whether variable $j$ is important for accurate prediction.

The models we consider predict $y$ by $\hat y = f(\bsx)\in\real$. Importance of a variable can be `local', meaning that we want to explain $\hat y_t=f(\bsx_t)$ for a target point $t\in1{:}n$. It can also be global, meaning that we want to quantify the importance of variables in aggregate for a set of points.

To measure the importance of $x_j$ to $f(\bsx)$ it is usual to change $x_j$, so $\bsx$ becomes $\tilde\bsx$, and to record the corresponding change $f(\tilde\bsx)-f(\bsx)$. There are an enormous number of ways to do this. We have to choose the value or values of $x_j$ to start from, and the value or values to change it to. Then we have to decide whether or not to also change $x_{j'}$ for $j'\ne j$ while we are changing $x_j$, and if we choose to change $x_{j'}$ we must specify how. When multiple changes are under consideration, we must find a way to aggregate them and then compare or rank the aggregate measures for variables $j\in1{:}d$.

Here are a few concrete examples from among many choices. Classical sensitivity analysis based on $|\partial f(\bsx_0)/\partial x_j|$ starts with a default point $\bsx_0\in\real^d$ and considers infinitesimally small changes there, assuming that $f$ is differentiable and the $x_j$ have been properly scaled relative to each other. The method of \citet{morr:1991} makes small (local) changes of size $\Delta_j$ to $x_j$ from a fine grid of starting points for $x_j$ with $x_{j'}$ for $j'\ne j$ sampled independently and uniformly over their values. It looks at the mean, standard deviation and even the whole cumulative distribution function of the resulting $f(\tilde\bsx)-f(\bsx)$ values. \citet{camp:cari:salt:2007} propose to aggregate $|f(\tilde\bsx)-f(\bsx)|$ instead. Global sensitivity analysis \citep{raza:etal:2021}, about which we say more below, considers not just local changes but all possible changes from random $\bsx$ to random $\tilde\bsx$ with some but not all $\tilde x_j=x_j$, aggregating the changes by second moments. The variable importance measure in random forests of \citet{brei:2001} makes a random permutation of $x_{1j},\dots,x_{nj}$ with respect to the other components of $\bsx_1,\dots,\bsx_n$ and records how prediction accuracy changes in response to permuting the $j$th variable.

There is not room to survey all variable importance methods and we must therefore leave many of them out and focus on the ones most closely related to our proposed method. \citet{wei:lu:song:2015} provide a comprehensive survey of variable importance measures in statistics, organized around a taxonomy. They cite 197 references of which 24 are themselves surveys. The global sensitivity analysis literature is focused on climate models, computer aided design models and similar tools that often have a strong physical sciences emphasis. In addition to the recent survey of \citet{raza:etal:2021}, with over 350 references, there are also textbooks of \citet{salt:etal:2008} and \citet{dave:gamb:ioos:prie:2021}. \citet{moln:2018} surveys the explainable AI (XAI) literature. Two prominent methods discussed there are LIME \citep{ribe:etal:2016} and SHAP \citep{lund:lee:2017}. A survey of methods based on Shapley value appears in \citet{sund:najm:2019}, which appeared as we were completing \citet{mase:owen:seil:2019}. %

\subsection{Attribution versus prediction}
Our objective is to study $f$ to explain it, and this is easier than studying a ground truth quantity like $\e(Y\giv \bsx)$ because $f$ is in our computer while $\e(Y\giv \bsx)$ must ordinarily be estimated by gathering real-world data. A question about what changing $x_j$ to $\tilde x_j$ would do to $f(\cdot)$ can be directly answered by computing $f(\bsx_{-j}{:}\tilde\bsx_j)-f(\bsx)$ for any setting $\bsx_{-j}$ of the other $d-1$ variables. Problems about causal inference when directed towards studying which variables cause $f$ to change are amenable to direct computational solutions. On the other hand, explaining why $f(\bsx)=1$ instead of $f(\bsx)=0$ is harder. There could be several $j\in1{:}d$ that would have given rise to $f(\bsx_{-j}{:}\tilde \bsx_j)=0$ had we changed $x_j$ to some other value $\tilde x_j$. In some problems every $j$ could have changed $f$. We might also have to consider the effects of changing more than one component of $\bsx$ at a time. We then have to consider which changes to one or more components $\bsx$ are most worthy of consideration.

The variable importance problem is one of studying `causes of effects' and not `effects of causes'. This distinction has been made repeatedly by Paul Holland; see, for example, \citet{holl:1986,holl:1988} citing \citet{mill:1851} and other philosophers. Holland's view is that statistical analysis is better suited to studying effects of causes. 
There is an excellent discussion of the difference between causes of effects and effects of causes in \citet{dawi:musi:2021} that also considers legal uses, such as a case about attributing a person's development of type II diabetes to past consumption of Lipitor. They consider at length the difficulties of
using potential outcomes \citep{rubi:1974} or structural causal models
\citep{pear:2009} in settings where there is only one variable being
considered as the possible cause of some effect.  In Section
16, they mention briefly that the setting with multiple
putative causes brings further difficulties.  
That is precisely the problem one faces in variable importance
settings.  The following quote is from their conclusions:
\begin{quotation}
We have presented a thorough account of a number of ways in which the statistical problems of effects of causes (EoC) and of causes of effects (CoE) have been formulated. Although most treatments of statistical causality use essentially identical tools to address both these problem areas, we consider that this is inappropriate. 
\end{quotation}

Identifying causes of effects is a problem of attribution. We seek to measure the extent to which different variables $x_j$ have contributed to the value of $f(\bsx)$. The big differences between attribution methods come down to their different definitions, which have different strengths and weaknesses. We cannot expect a `bakeoff' where algorithms compete to estimate some known true attributions to settle a discussion about which definition is most appropriate. This means that the common task framework \citep{dono:2019} that has been so effective in improving prediction methods has limited utility for attribution.
A second issue that Holland raises about `causes of effects' is that the cause we identify might itself be an effect of some other cause. In causal inference, we can avoid adjusting for variables whose values became known after an intervention. For attribution, it is not so clear how to separate proximate causes from some prior causes. We might also have to think of even earlier ultimate causes of those prior causes.

We are familiar with the problem of attribution in real life. If a candidate lost an election by $10{,}000$ votes and some issue caused an unfavorable vote swing of $11{,}000$ votes, then we could at first reasonably attribute her loss to that issue. However, there may be ten such issues that individually or in combination brought a swing of over $10{,}000$ votes. It is then less clear that the first issue is `the reason'.
The attribution issue has been intensely studied in advertising. In media mix modeling (MMM), users are told what percentage of their sales to attribute to each advertising channel; see \citet{chan:perr:2017} for a discussion of challenges in MMM. In online advertising, Shapley values are now being used instead of simply attributing a sale or other customer conversion to the last click prior to that event \citep{berm:2018}. Shapley values are also being used in financial profit-and-loss attribution \citep{moeh:boyd:ang:2021}.

One maxim in causal inference is that there is `no causation without manipulation'; see, for instance, \newt{\citet{holl:1986}}. Under this maxim, if we cannot in principal intervene to change a variable then we cannot make claims about its causal effect, \newt{at least not via the potential outcomes framework where both a treatment
and an alternative must be possible.} \newt{Judea Pearl disagrees
with this maxim.  For instance
\cite{bollen2013eight} remark that this principal would
lead to a conclusion that the moon does not cause tides.}
In problems of attribution, we cannot avoid discussing the importance of variables, such as a specific person's birth year or ancestry, that an investigator cannot intervene to change.

\subsection{Challenges to variable importance: interaction and dependence}
Now we turn to quantifying the effect of $x_j$ on $f(\bsx)$. The setting is simplest for an additive model of independent random inputs.
Suppose that
\begin{align}\label{eq:fadditive}f(\bsx) = \sum_{j=1}^df_j(x_j)
\end{align}
where $x_j$ has support $\cx_j$ and $x_1,\dots,x_d$ are independent random quantities with $f_j:\cx_j\to\real$.
Global importance measures can then be defined in terms of semi-norms of the $f_j$. We could choose $\var(f_j(x_j))$ or $\var(f_j(x_j))^{1/2}$ or $\sup_{x_j\in\cx_j}f_j(x_j)-\inf_{x_j\in\cx_j}f_j(x_j)$. If $\cx_j\subseteq\real$ and $f_j$ has derivative $f_j'$, then we could choose $\sup_{x_j\in\cx_j}|f'(x_j)|$ or the total variation $\int_{\cx_j}|f'(x_j)|\rd x_j$ or $\e(|f'(x_j)|)$. Semi-norms make sense because replacing $f_j(x_j)$ by $f_j(x_j)+c$ and $f_{j'}(x_{j'})$ by $f_{j'}(x_{j'})-c$ does not reasonably change the importance of $x_j$ and $x_{j'}$. Implicit in each of these \newt{measures} is a choice of how to combine the effects of changes in $f_j$ and hence in $f$ over pairs $x_j,\tilde x_j\in\cx_j$. These \newt{various semi-norms} quantify importance in different ways, but \newt{the setting of equation~\eqref{eq:fadditive}} is the one where it is most straightforward to choose a definition for a given problem \newt{for global and also local variable importance.} The effect of changing $x_j$ is unrelated to the value of $x_{j'}$ and given a value for $x_j$ we have no reason to direct our attention about $x_{j'}$ to any specific part of $\cx_{j'}$.

Variable importance becomes much more difficult when the additivity condition~\eqref{eq:fadditive} does not hold. In that case the variables have interactions, and the importance of those interactions has to be shared somehow among the variables that contribute. A second complication is that the distribution of $\bsx$ strongly affects which variable is important. To illustrate this effect, note that according to the CDC \url{https://www.cdc.gov/measles/about/faqs.html} about 3\% of people immunized against measles will get measles if they are exposed to it. Let's assume that anybody not exposed does not get measles and also, since it is very contagious, that 90\% of people not immunized will get it if exposed. Table~\ref{tab:cdcmeasles} illustrates how an interaction can complicate the relative importance of variables. If almost everybody is in the upper left \newt{cell}, then exposure is the more important variable. If almost everybody is in the lower right \newt{cell} then immunity is the more important variable. \newt{Note that this second complication from interactions is present already in
the setting of independent input variables.} For a thorough account of the phenomenon of interaction, see \citet{cox:1984} and \citet{dego:cox:2007}.

\newt{While} interactions can be conveniently handled by Sobol' indices based on the ANOVA decomposition as we will see in Section~\ref{sec:techniques}, dependence among the components of $\bsx$ is far harder to deal with. In that case, the reasonable changes to $x_j$ can depend on all components in $\bsx_{-j}$ and their combinations. The Boston housing data illustration in Figure~\ref{fig:dontpermute}
shows a setting with two variables where at least one must be near a default value. An extreme version of this problem is given in \citet{owen:prie:2017} where $\bsx\in\{0,1\}^2$ but the only possible values of $\bsx$ are $\{0,0\}$, $\{0,1\}$ and $\{1,0\}$. If $\bsx=\{0,1\}$ then it is not meaningful to change $x_1$ while leaving $x_2$ unchanged, so such a change cannot contribute to the importance of variable $1$. Higher dimensional patterns will be much harder to detect. 

\newt{The challenges brought by dependence are
not easy to handle via the ANOVA}.
See \citet{chas:gamb:prie:2012} and \citet{owen:prie:2017}. 
When there are dependencies among variables, we can turn from ANOVA to the Shapley value that we describe in Section \ref{sec:shapley4importance}.
Shapley value opens up some conceptually attractive solutions that nonetheless have computational challenges such as estimating an exponentially large number of conditional expectations.

\begin{table}[t]
    \centering
    \begin{tabular}{lcc}
    \toprule
      $\Pr(Y\giv\bsx)$ &Not Immune & Immune \\
      \midrule
Not exposed & 0.00 & 0.00\\
 Exposed & 0.90 & 0.03\\
 \bottomrule
    \end{tabular}
    \caption{\label{tab:cdcmeasles} The lower right hand \newt{cell} has the CDC's estimate of the probability that a person immunized against measles will catch it, if exposed. The other three numbers are hypothetical. These variables interact. As noted in the text either exposure or immunity could be more important depending on how the $\bsx=(\text{exposed},\text{immune})$ values are distributed.}
    
\end{table}

Our approach of studying changes to $f$ does not capture every variable importance measure in widespread use. For instance the importance of predictors in spike-and-slab regression models is often measured by the posterior probability that $\beta_j\ne0$ which is not directly obtainable from $f(\bsx)=\wh\e(y\giv \bsx)$ \citep{scot:vari:2014}. Such measures provide a useful aggregate for the role of $x_j$ but do not easily explain local decisions.
Similarly, counts of the number of times a variable is used to define a split in a tree or forest model do not have a direct interpretation in terms of the value of $f$.

\section{ANOVA, Sobol' indices, and Shapley value}\label{sec:techniques}

In this section we describe some of the tools we need in order to present the cohort Shapley measure. These are general analysis of variance (ANOVA) decomposition, Sobol' indices, and Shapley value. %

\subsection{General ANOVA}
When the components of $\bsx$ are independent and $\e(f(\bsx)^2)<\infty$, there is a convenient analysis of variance (ANOVA) decomposition of $f$. This ANOVA can be used to define variance components and subsequently Sobol' indices. Sobol' indices provide a very powerful way to quantify global importance of variables. They do not provide local explanations, and, as we will see below, they have difficulty with dependent data settings.

The familiar ANOVA used in experimental design applies to tabular data defined in terms of categorical $x_j$. It goes back to \citet{fish:mack:1923}.
A generalization to $\bsx\sim\dunif[0,1]^d$ was used by \citet{hoef:1948}, \citet{sobo:1969} and \citet{efro:stei:1981} for U-statistics, integration and the jackknife, respectively. The ANOVA decomposes $\sigma^2=\var(f(\bsx))$ into components for each $u\subseteq1{:}d$. For $d=\infty$ see \citet{lss}.

\newt{To understand the generalized ANOVA, we recall
how the orginal one works. We are given $f(\bsx)$ where
now $x_j$ takes $k_j$ levels and we have all
$N=\prod_{j=1}^dk_j$ possible evaluations of $f$. The $x_j$ are then independent random variables under equal weighting of those $N$ values. We first compute a grand mean $\mu$ averaging
all the values $f(\bsx_1),\dots,f(\bsx_N)$. Next, to define 
the main  effect of variable $j$, we subtract $\mu$ from each 
$f(\bsx_i)$ and at each level of $x_j$ we average those
differences over $N/k_j$ settings of the other $d-1$ variables.
The idea is that only differences merit an explanation.  For
an interaction between variables $j$ and $j'$, we subtract the
grand mean and both main effects from $f(\bsx_i)$ and for
each of the $k_jk_{j'}$ combinations of $x_j$ and $x_{j'}$
we average the results over all $N/(k_jk_{j'})$ settings
of the other $d-2$ variables. The general rule is like this:
we don't attribute to $\bsx_u$  what can be explained by
$\bsx_v$ for $v\subsetneq u$, then to get a function of
$\bsx_u$ we average over $\bsx_{-u}$.}

The general ANOVA replaces averages by expectations.
It begins with a grand mean function
$f_\emptyset(\bsx)$ which is everywhere equal to
the constant $\mu\equiv\e( f(\bsx))$. The main effect for variable $j$ is
the function
$$f_{\{j\}}(\bsx) = \e( f(\bsx)-\mu\giv \bsx_j)=\e(f(\bsx)\giv \bsx_j)-\mu.$$
For a subset of variables $u\subseteq1{:}d$, we define
\begin{align}\label{eq:deffu}
f_u(\bsx) =\e\Bigl(
f(\bsx)-\sum_{v\subsetneq u}f_v(\bsx) \!\bigm|\! \bsx_{u}
\Bigr)=\e(
f(\bsx)\mid \bsx_{u})
-\sum_{v\subsetneq u}f_v(\bsx).
\end{align}
The first expression in~\eqref{eq:deffu} shows that we do not attribute to $\bsx_u$ what can be explained by $\bsx_v$ for any proper subset $v$ of $u$. The second expression follows because $f_v$ is not random given $\bsx_u$.  While $f_u$ is defined 
on the whole domain of $\bsx$ its value only depends on $\bsx_u$.

When $|u|>1$, the effect $f_u$ is a $|u|$-fold interaction.
The effect $f_u$ has variance component $\sigma^2_u =\var(f_u(\bsx))$. We easily find that
$$
f(\bsx) = \sum_{u\subseteq1{:}d}f_u(\bsx)\quad\text{and}\quad \sigma^2=\sum_{u\subseteq1{:}d}\sigma^2_u.
$$

\subsection{Sobol' indices}
Sobol' indices \citep{sobo:1993} measure the importance of the subvector $\bsx_u$ not just individual variables $x_j$. There are two versions, the closed index and the total index, respectively,
$$
\ult^2_u = \sum_{v\subseteq u}\sigma^2_v\quad\text{and}\quad\olt^2_u = \sum_{v:v\cap u\ne\emptyset}\sigma^2_v.
$$
These are usually divided by $\sigma^2$ to give a proportion of variance explained. We easily see that $0\le \ult^2_u \le \olt^2_u = \sigma^2-\ult^2_{-u}$.
The total index $\olt^2_u$ includes interactions between $\bsx_u$ and $\bsx_{-u}$ that the closed index $\ult^2_u$ excludes. If $\ult^2_u$ is relatively large then we interpret $\bsx_u$ as important while if $\olt^2_u$ is small then $\bsx_u$ can be interpreted as unimportant.
\newt{One can show that $\ult^2_u =\var( \e(f(\bsx)\giv \bsx_u))$ and $\olt^2_u =
\e(\var(f(\bsx)\giv \bsx_{-u}))$.}

Sobol' indices have been intensely studied in the global sensitivity analysis literature. They are very conveniently estimated by sampling methods based on identities like
$$
\ult^2_u = \e\bigl( f(\bsx)f(\bsx_u{:}\bsz_{-u})\bigr)-\mu^2
\quad\text{and}\quad
\olt^2_u = \frac12\e\bigl((f(\bsx)-f(\bsx_{-u}{:}\bsz_{u}))^2\bigr)
$$
due to \citet{sobo:1993} and \citet{jans:1999}, respectively. 

Sobol' indices provide a very good global measure of how important a variable or even a set of variables is. They are not aimed at variable importance at a specific point $\bsx$, \newt{so they are not suited
to local importance.}

\newt{We mentioned earlier that the ANOVA does not
extend well to dependent data settings. Since
Sobol' indices are based on the ANOVA, this causes
difficulties for them.}
The ANOVA for dependent data has been considered by \citet{chas:gamb:prie:2012} building on work of \citet{hook:2007} who builds on that of \citet{ston:1994}.
\newt{One of the challenges is that some ways to
define ANOVA lead to negative analogues of variance components for dependent data.}
A key condition \newt{in some of the definitions}
is that the density $p(\bsx)$ must be bounded below by $c\times p_u(\bsx_u)\times p_{-u}(\bsx_{-u})$ for some $c>0$ and all $\bsx$. This rules out distributions with `holes' where $p(\bsx)=0$ on a rectangle within the support of $p$. It also rules out multivariate Gaussian distributions with nonzero correlations. 

\newt{The two aforementioned problems with Sobol' indices
do not apply to Shapley value which we consider next.}

\subsection{Shapley value}\label{sec:shapval}
The Shapley value \citep{shap:1953} is used in cooperative game theory to define a fair allocation of rewards to team members who have jointly produced some value. It has seen many uses in defining variable importance measures. See \citet{sund:najm:2019} and \citet{moln:2018} for surveys of uses in XAI and \citet{song:nels:stau:2016} in global sensitivity analysis. The analogy is to view the variables $x_j$ in a model $f(\bsx)$ as team members cooperating on a goal such as predicting~$y$.

In Shapley's formulation, there is a set $1{:}d$ of entities. Together they produce a value that we denote by $\nu(1{:}d)$. We suppose that we can find the value $\nu(u)$ that would have been produced by any $u\subseteq1{:}d$. It is customary to assume that $\nu(\emptyset)=0$. Shapley proposes four quite reasonable axioms on the basis of which there is then a unique value $\phi_j=\phi_j[\val]$ that should be attributed to player $j$ on the team. A key quantity in Shapley's formulation is the incremental value that would be brought to a team $u$ if player $j\not\in u$ were to join it. This incremental value of player $j$ given that players in $u$ are already on the team is
$$
\nu(j\giv u) = \nu( u\cup\{j\})-\nu(u).
$$

Shapley's four axioms are:
\begin{compactenum}[\qquad \bf 1)]
\item Efficiency: $\sum_{j=1}^d\phi_j=\nu(1{:}d)$.
\item Dummy: If $\nu(j\giv u)=0$ whenever $j\not\in u$, then $\phi_j=0$.
\item Symmetry: If $\nu(j\giv u)=\nu(j'\giv u)$ whenever $j,j'\not\in u$ then $\phi_j=\phi_{j'}$.
\item Additivity: For games $\nu$ and $\nu'$, $\phi_j[\nu+\nu']=\phi_j[\nu]+\phi_j[\nu']$.
\end{compactenum}

The second and third axioms are compelling. \citet{moln:2018} suggests that the first axiom might even be a requirement in settings where one is required by law to provide an explanation. See also \citet{doshi2017accountability}. The fourth axiom is more debatable but is usually included in variable importance applications and we will use it. Under these four axioms there is a unique solution
\begin{align}\label{eq:shapleyval}
\phi_j = \frac1d\sum_{u\subseteq -j}{d-1\choose |u|}^{-1}\nu(j\giv u).
\end{align}

An intuitive explanation of~\eqref{eq:shapleyval} is as follows: Suppose that we build a team from $\emptyset$ to $1{:}d$ in one of $d!$ possible orders and track the incremental value that arises at the moment when player $j$ joins this team. Then $\phi_j$ is the average of those $d!$ incremental values. Actually averaging all permutations would have a cost that is $O(d!)$. The direct calculation of~\eqref{eq:shapleyval} depends on only $d2^{d-1}$ incremental values: each of $2^d$ sets $u$ has $d$ neighboring sets that differ by inclusion or deletion of exactly one of the $d$ team members. Every pair $(u,j)$ with $u\subset1{:}d$ and $j\not\in u$ is counted twice this way.

\newt{For illustration, suppose that $d=3$ as in our running
example.
Then there are $6$ permutations of the three variables. Variable
$1$ enters first in two of them, following $\emptyset$.
It enters last in two of them, following $\{2,3\}$.
It enters once after just $\{2\}$ is present and once after
just $\{3\}$ is present.
Therefore
\begin{align}\label{eq:phi1ford3}
\phi_1 &= \frac16\Bigl(
2\bigl[ \nu(\{1\})-\nu(\emptyset)]
+2\bigl[ \nu(\{1,2,3\})-\nu(\{2,3\})]
+\bigl[\nu(\{1,2\})-\nu(\{2\})]+\bigl[\nu(\{1,3\})-\nu(\{3\})]
\Bigr)
\end{align}
}
and the other $\phi_j$ are similar.

The Shapley values $\phi_j$ are unchanged if we add a constant $c$ to all of the values $\nu(u)$. This follows because $c$ cancels out of the incremental values. As remarked above, it is common to assume that $\nu(\emptyset)=0$. Sometimes it is simpler to define a value function in a way that does not force $\nu(\emptyset)=0$. In such cases the Shapley values attribute the incremental value $\nu(1{:}d)-\nu(\emptyset)$ of the whole team to the members of that team.

The cost to compute Shapley values exactly grows exponentially in $d$. The representation via permutations opens up a natural Monte Carlo strategy of using an average over a sample of randomly generated permutations. Monte Carlo is routinely used in settings such as integration where an exact computation would have infinite cost. Its effectiveness depends on the variance of the incremental values and not on the number $d2^{d-1}$ of incremental values needed for exact calculation. \citet{michalak2013efficient} and \citet{mitchell2022sampling} describe some value functions where they find that Monte Carlo is inefficient. For cohort Shapley, Monte Carlo would be unfavorable if only a vanishing fraction of the incremental values $\val(j\giv u)$ were non-negligible. We consider this unlikely and have not seen it in our experience. 

The integrated gradients method of \citet{sundararajan2017axiomatic} is based on the Aumann-Shapley value of \citet{auma:shap:1974}. Where the Shapley value averages difference over paths on the edges of $\{0,1\}^d$, integrated gradients average a gradient over the diagonal of $[0,1]^d$ which can be much faster to compute.

\section{Using Shapley for importance}\label{sec:shapley4importance}

In this section, we show how Shapley values have been used to study variable importance. Then we present the cohort Shapley algorithm that only uses actually observed data for variable importance. We also include a discussion of some tradeoffs and choices that come up when using Shapley values for variable importance.
There are many different ways to formulate a game with which to apply the Shapley formulations. The axioms have considerable logical force, conditional on the value function being a reasonable one. As a result, the important differences among methods based on Shapley values stem from the choice of value function.

\subsection{Some past uses of Shapley value}

\citet{lind:mere:gold:1980} propose a measure of the importance of including a variable in a linear regression model. Their proposal is equivalent to using the Shapley value with $\nu(u)$ is \newt{proportion of the variance explained by a linear model (usually denoted $R^2$) using
the variables $\bsx_u$.}
\citet{stru:kono:2010} used Shapley value to explain the value of $f(\bsx_t)$ in a local explanation for target observation $t$. \newt{In their equation (1),} they use the value function
$$
\nu(u) = \e( f(\bsx_{t,u}{:}\bsx_{-u}))-\e(f(\bsx))
$$
\newt{where the first 
expectation is taken with respect
to independent discrete random 
variables $\bsx_j$ for $j\not\in u$
and the second one has all $\bsx_j$
from independent discrete distributions.
Those discrete distributions could be
the empirical marginal distributions.
They also consider random sampling approximation
that allows continuous distributions.}

For $\bsx$ with independent components, \citet{owen:2014} takes $\nu(u)$ to be the variance explained by $\bsx_u$, $\var( \e(f(\bsx)\giv \bsx_u))=\ult^2_u$. The result is that
$$
\phi_j = \sum_{u\subseteq1{:}d}
\frac{1\{j\in u\}\sigma^2_u}{|u|}.
$$
Every variance component $\sigma^2_u$ is shared equally among the $|u|$ variables that contribute to it. It follows that the easily estimated Sobol' indices bracket the Shapley value:
$$
\ult^2_j\le \phi_j\le \olt^2_j.$$
\citet{plis:rabi:borg:2021} note that the upper bound can be improved to $(\olt^2_j+\ult^2_j)/2$.

\citet{song:nels:stau:2016} propose the use of Shapley value with $\val(u)=\var(\e(f(\bsx)\giv \bsx_u))$ for a setting where components of $\bsx$ may be dependent. They present a computational algorithm based on permutations and they show that the same Shapley values arise using $\val(u)=\e(\var(f(\bsx)\giv \bsx_{-u}))$. \citet{owen:prie:2017} describe how Shapley values solve the conceptual problems that dependence causes for Sobol' indices, though computational challenges remain.

Baseline Shapley (see \citet{sund:najm:2019}) explains $f(\bsx_t)-f(\bsx_b)$ where $t$ is a target point and $\bsx_b$ is a `baseline' vector of predictors. It could be made of default values or it could be $\bar\bsx =(1/n)\sum_{i=1}^n\bsx_i$, although some components of $\bar\bsx$ might not be possible feature values. In baseline Shapley, the value function is $\val(u)=f(\bsx_{t,u}{:}\bsx_{b,-u})$ and unobserved hybrid values $\bsx_{t,u}{:}\bsx_{b,-u}$ are to be expected.

Many of the alternatives to baseline Shapley use a `conditional expectation Shapley' with 
$$\val(u) =\e( f(\bsx_t)\giv \bsx_{t,u})$$
differing according to which distribution of $\bsx_{t,-u}$ given $\bsx_{t,u}$ they use to define the expectation. For instance the above mentioned choice of \citet{stru:kono:2010} is of this type assuming independent predictors.
The quantitative input influence of \citet{datt:2016} takes $\val(u)=\e(f(\bsx_u{:}\bsz_{-u})\giv\bsx_u)$ under a sampling model where $\bsz_{-u}$ has the empirical marginal distribution of $\bsx_{-u}$. That will generally produce unobserved hybrid points. Kernel SHAP \citep{lund:lee:2017}, similarly uses an independence assumption on inputs to compute a conditional Shapley value.

The TreeSHAP method \citep{lund:2020} is designed to be especially computationally efficient on tree structured models but can only be used on such models. If the prediction is a weighted sum of tree predictions then, using additivity of the Shapley value one can take a weighted sum of Shapley values of single trees. For a single tree, the value is $\e(f(\bsx)\giv\bsx_u)$ where the distribution of $f(\bsx_{-u})$ given $\bsx_u$ is uniform over all the values of $f(\bsx_i)$ in all the leaves of the tree that are compatible with $\bsx_u$. It only uses observed data and it cannot attribute importance to variables that are not used by the model.

\citet{frye:2021} share our concerns about using impossible data. They use a variational autoencoder (VAE) to represent the $d$ predictors as being scattered around a lower dimensional manifold. They compute $\e(f(\bsx)\giv \bsx_u)$ using their model to define the distribution of $\bsx_{-u}$ given $\bsx_u$, but not every point $\bsx$ they use will have been observed, \newt{so whether those are reasonable depends on performance of the chosen VAE.}
Similarly, \citet{aas:2019} propose a conditional kernel SHAP that uses weights from an ellipsoidally shaped kernel based on a Mahalanobis distance between target and reference points using the sample covariance. They need a scaling parameter for each subset of input variables. Their kernel puts positive weight on unobserved variable combinations.

Finally, we note that \citet{ghor:zou:2019} use Shapley value to measure the value of individual data points in a model. They compare this to leave-one-out methods and leverage scores and find that it helps to identify outliers and corrupted data. For instance, data points with a negative Shapley value for model accuracy are worthy of inspection; \newt{perhaps the response value was mislabeled or there is some other error in that data.}

Some more uses of Shapley value in XAI are surveyed in \citet{sund:najm:2019}, \citet{moln:2018} and \citet{kuma:2020}.

\subsection{Cohort Shapley}

The cohort Shapley method quantifies variable importance using only actually observed data values. Our description here is based on \citet{mase:owen:seil:2019}. 

For target  $t$ and variable $j$ we define a function $s_{t,j}$ where $s_{t,j}(\bsx_{ij})=1$ if $\bsx_{ij}$ is similar to $\bsx_{tj}$ and is zero otherwise. It must always be true that $s_{t,j}(\bsx_{tj})=1$. For binary variables, similarity is just whether
$\bsx_{ij}=\bsx_{tj}$. For real valued features $j$, similarity could be that $|\bsx_{ij}-\bsx_{tj}|\le\delta$ or $|\bsx_{ij}-\bsx_{tj}|\le \delta |\bsx_{tj}|$ for some $\delta>0$, or we could discretize $\bsx_{ij}$ values into a set of ranges and take observations to be similar if they are in the same range. This last choice makes similarity transitive, which the other choices do not impose.

For observation $t$ and $u\subseteq1{:}d$, define the cohort
\begin{align}\label{eq:defcohort}
C_t(u) = \{ i\in 1{:}n\mid s_{t,j}(\bsx_{ij})=1,\ \text{all $j\in u$}\},
\end{align}
with $C_t(\emptyset)=1{:}n$ by convention.
The value function in cohort Shapley is
$$
\val(u) =\frac1{|C_t(u)|}\sum_{i\in C_t(u)}f(\bsx_i),
$$
for local explanation for observation $t$. 
These values are all well defined because $t\in C_t(u)$ always holds.
In the random permutation framework, we begin with the cohort $C_t(\emptyset)$ of all observations whose mean is then the global mean $(1/n)\sum_{i=1}^nf(\bsx_i)$. We then refine the cohort, imposing similarity with respect to all $d$ variables in order. An important variable is one that brings a relatively large change to the cohort mean when we refine on it. The cohort Shapley values $\phi_j$ explain the difference between the mean of $f(\bsx_i)$ over the fully refined cohort $C_t(1{:}d)$ and the global mean.  If all input variables have been binned, then cohort Shapley becomes a conditional expectation Shapley defined in terms of the binned variables.

In applying cohort Shapley, one is forced to define what it means for $\bsx_{ij}$ to be similar to $\bsx_{tj}$. This is difficult for continuous variables. Modest numbers of levels are usually recommended in similar settings that require `binning' a continuous variable.
\citet{gelm:park:2009} use means over the largest and smallest of just three groups to approximate a regression slope with small loss of efficiency. \citet{coch:1968} finds diminishing returns to using more than five strata in sampling problems. Cochran's results are based on linear regression and nonlinear relationships would reasonably require a few more strata. For a modestly nonlinear setting like propensity scores for logistic regression, \citet{neuh:thie:ruxt:2018} find diminishing returns at around ten strata. 

While binning a variable can seem arbitrary we think that it is more transparent than employing yet another black box, such as a \newt{VAE}, to approximate empirical conditional distributions.  

We can understand TreeSHAP in terms of cohorts. For a single tree it uses cohorts defined by a union of leaves of that tree. Such a union of leaves need not define a spatially connected set when $\cx =\real^d$. For aggregates of multiple trees the notion of similarity will generally differ between the trees in that aggregate.  Our concern with TreeSHAP is that it uses a notion of variable similarity defined in part by the response values it is fitting.  This makes it harder to interpret or explain the underlying similarity concept.  
In a high stakes setting, such as whether a given algorithm was unfair to one or more people,
we would be interested to know who those 
people were compared to.  It is also not able to attribute importance to a variable that the model does not use.

Categorical variables with many levels may need to be coarsened similarly to continuous variables. If all of the variables are categorical or have been made categorical by stratification then cohort Shapley matches conditional expectation Shapley based on the empirical distribution of the data.

\subsection{Running example revisited}

Table~\ref{tab:cohorts} shows the cohorts of target observation $1$ 
for our running
example.  Potential observation 7 is unobserved in one of
the data distributions and so it is removed from the two
cohorts that it would have appeared in had all $8$ $\bsx$ values
been observed. That table also records the cohort means for our
two distributions and two example functions\newt{, $f_1(\bsx) = x_1 - 2 x_2$ and $f_2(\bsx) = x_1 x_2 (1 - x_3)$}.

\begin{table}
\centering
\begin{tabular}{ccccccc}
\toprule
Set $u$ & Cohort & Cohort$_*$ & $\bar f_1$ &$\bar f_{1*}$ & $\bar f_2$ & $\bar f_{2*}$\\
\midrule
$\emptyset$ & $\{1,2,3,4,5,6,{\bf7},8\}$ & $\{1,2,3,4,5,6,8\}$&$-1/2$&$-3/7$&$1/8$ &0\\
$\{1\}$ & $\{1,2,3,4\}$ & $\{1,2,3,4\}$ &$-1$&$-1$&0&0\\
$\{2\}$ & $\{1,2,5,6\}$ & $\{1,2,5,6\}$&$\phm1/2$&$\phm1/2$&0&0\\
$\{3\}$ & $\{1,3,5,{\bf7}\}$ & $\{1,3,5\}$&$-1/2$&$-1/3$&$1/4$&$0$\\
$\{1,2\}$ & $\{1,2\}$ & $\{1,2\}$ &$\phm0$ &$\phm0$&0&0\\
$\{1,3\}$ & $\{1,3\}$ & $\{1,3\}$ &$-1$ & $-1$&0&0\\
$\{2,3\}$ & $\{1,5\}$ & $\{1,5\}$ &$\phm1/2$ & $\phm1/2$ &0 &0\\
$\{1,2,3\}$ & $\{1\}$ & $\{1\}$ & $\phm0$ & $\phm0$&0&0\\
\bottomrule
\end{tabular}
\caption{\label{tab:cohorts}
For the example setting in Table~\ref{tab:runningexample},
this table shows the cohorts of target
observation $t=1$ for all $8$ subsets of 
variables $\{1,2,3\}$. There is a column for the
distribution labeled $P$ and one for the distribution
labeled $P_*$. Impossible combination $7$ does not
appear in the second set of cohorts. The final four
columns show cohort means for two functions and
the two distributions.
}
\end{table}

\newt{
Table~\ref{tab:csvalues} shows cohort Shapley values for our
running example.  For independent data and $f_1(\bsx)=x_1-2x_2$
the average value of $f(\bsx_i)$ over all data (cohort $C_1(\emptyset)$)
is $\nu(\emptyset)=-1/2$. For target observation $t=1$ 
and cohort $C_1(\{1,2,3\})=\{1\}$ we get $\nu({\{1,2,3\}})=0$.
Therefore we have to explain why $f(\bsx_t)$ is $1/2$ unit
above average. There are six equally weighted paths from $\emptyset$
to $\{1,2,3\}$. 
As described at equation~\eqref{eq:phi1ford3}, we get
\begin{align*}
\phi_1 
 &= \frac16\Bigl(
2\bigl[ -1/2]+2\bigl[ -1/2]
+\bigl[-1/2]+\bigl[-1/2]
\Bigr)=-\frac12.
\end{align*}
Similarly, $\phi_2=1$ and $\phi_3=0$.
So the target observation $t=1$ has $f$ that is $1/2$ unit above average
and the cohort Shapley attributions are that this
arises from a $-1/2$ effect of having $x_1=0$
offset by a $+1$ effect of having $x_2=0$ while
$x_3$ does nothing.  We see that $\phi_j$ takes account
of not just the sign of the coefficient of $x_j$
but also whether $x_{tj}$ is at the low or high value
of $x_j$.}

\begin{table}[b]\centering %
\begin{tabular}{lrrrrcccc}
\toprule
Dist & $f$ & $\phi_1$ & $\phi_2$ & $\phi_3$\\
\midrule
$P$ & $f_1$ & $-0.5$ & $\phm1$ & $\phm0$\\
$P_*$ & $f_1$ & $-0.55$ & $0.95$&$\phm0.03$\\
$P$ & $f_2$ & $-1/12$ & $-1/12$ & $\phm1/24$\\
$P_*$ & $f_2$ &$\phm0$&$\phm0$&$\phm0$\\
\bottomrule
\end{tabular}
\caption{\label{tab:csvalues}
This table shows the cohort Shapley values for variables
$1$ to $3$ in our running example under both example distributions
for (scaled) versions of both functions.
The exact values for $P_*$ and $f_1$ are
$-139/252$, $239/252$ and $8/252=2/63$.
}
\end{table}

\newt{
If $\bsx=(1,1,0)$ is impossible, as in distribution $P_*$, then that introduces
some amount of dependence between $x_3$ and the other variables.
Now $\phi_3\ne0$ for the target observation and the
linear function $f_1=x_1-2x_2$ that does not involve $x_3$.
For $f_2=x_1x_2(1-x_3)$ under the independence distribution $P$, 
the target observation
has $f_2(\bsx_t)=0$ while the average of $f_2$ is $1/8$.  The Shapley
attributions are $-1/12$ for both $x_1$ and $x_2$
and $+1/24$ for $x_3$ owing to the factor $1-x_3$.
Under $P_*$, $f_2$ is only nonzero at the impossible
combination and then all $\phi_j=0$.
}
\subsection{Tradeoffs and issues}

There are several difficult choices to consider in rating variable importance generally and some difficult choices specific to methods based on Shapley value. These stem from `importance' meaning different things in different contexts. \citet{kuma:2020} present several qualms about the use of Shapley value in any of its forms for variable importance.

One consequential decision is whether to attribute any importance to a variable $x_j$ that is not used at all in $f(\bsx)$. In a case like this, changes to $x_{tj}$ cannot possibly change the prediction $\hat y_t = f(\bsx_t)$ so it might seem unreasonable to attribute importance to $x_{tj}$. Suggesting that person $t$ change $x_{tj}$ to qualify for a loan would be bad advice. However there are settings in algorithmic fairness where a protected variable such as race or gender that is not used by $f(\cdot)$ might still play a role through its association with other variables that were used. In those applications, it is essential to consider the possibility that $x_j$ could be important. Fairness by unawareness is not considered reliable \citep{dwor:2012}. Cohort Shapley can be used to identify importance for variables not used by the function. To do that, one includes that variable in the set of predictors under study.  We illustrate this in Section~\ref{sec:compas}, in an example involving race.

A potential problem with Shapley values arises when two variables $x_j$ and $x_{j'}$ are very strongly related. This can cause them to `share their importance' which might make both of them appear unimportant. Or, as \citet{kuma:2020} note, if one variable denotes a protected category and the other does not, including both in the Shapley calculations could make the protected variable look less important than it really is.

\section{Algorithmic Fairness and the COMPAS data}\label{sec:compas}

Algorithmic fairness makes use of variable importance. For instance, it is unfair when a protected variable such as race or gender is improperly important in a decision that affects a person. 
In this section, we conduct a detailed investigation of the data from  \citet{angw:2016} regarding the fairness of the COMPAS tools and their use in recidivism prediction.
 The focal issue is whether COMPAS is unfair to Black defendants, and if so, by how much and to which defendants?
The results in this section were presented earlier in \citet{fairnessshapley}, which this article supercedes.

There are several aspects of cohort Shapley that make it very suitable to analyzing fairness of the COMPAS data:
\begin{compactenum}[\quad\bf1)]
\item
it is able to judge fairness with respect to race for an algorithm that was constructed without using race as a variable;
\item it is able to work without access to $f$, using instead its predictions, which in this instance were made by a proprietary algorithm that is not available to researchers;
\item the additivity property of Shapley value means that the Shapley values for residuals (actual versus predicted response) equal the corresponding differences between Shapley values for observations and their predictions;
\item the additivity property of Shapley value lets us aggregate values over a set of defendants, such as all Black women;
\item we can work with a data set that is not necessarily the one the algorithm was trained on;
\item because we have samples of $(\bsx_i,y_i,\hat y_i)$ for defendants $i$ we can use bootstrap sampling to get a measure of sampling uncertainty in our Shapley values for either individuals or aggregates of individuals.
\end{compactenum}
Cohort Shapley is the unique method from
Table~\ref{tab:affordances} to satisfy
condition 2 and one of only two that satisfy condition 1.
The uncertainty in condition 6 is distinct from uncertainty due to random sampling of permutations. In this analysis we are able to compute cohort Shapley exactly and the uncertainty arises from viewing the defendants as a sample from a larger population. This setting is simpler to bootstrap because the model was trained on a separate data set.
\subsection{Background on fairness}

Just as there are many ways to define variable
importance, there are multiple ways to define
what fairness means. Some of those definitions
are mutually incompatible and some of them differ
from legal definitions. Here we present just a few of the issues as a prelude to studying the COMPAS data.
See \citet{corb:goel:2018}, \citet{chou:roth:2018}, \citet{berk:2018}, and \citet{frie:etal:2019} for surveys.
We do not make assertions about which
definitions are preferable.

For $y,\hat y\in\{0,1\}$, let $n_{y\hat y}$ be
the number of observations with
$y_i=y$ and $\hat y_i=\hat y$.
These four counts and their derived properties can
be computed for any subset of the observations.
The false positive rate (FPR) is
$n_{01}/n_{0\sumdot}$, where a bullet indicates that we are summing over the levels of that index.
We ignore uninteresting corner cases such
as $n_{0\sumdot}=0$; when there are no
observations with $y=0$ then we have no interest
in the proportion of them with $\hat y=1$.
The false negative rate (FNR) is
$n_{10}/n_{1\sumdot}$.
The prevalence of the trait under study
is $p=n_{1\sumdot}/n_{\sumdot\sumdot}$.
The positive predictive value (PPV) is
$n_{11}/n_{\sumdot1}$.
As \citet[equation (2.6)]{chou:2017} notes,
these values satisfy
\begin{align}\label{eq:fprandp}
\fpr = \frac{p}{1-p}\frac{1-\ppv}{\ppv}(1-\fnr).
\end{align}
See also \citet{klei:mull:ragh:2016}.

Equation~\eqref{eq:fprandp} shows how some natural
definitions of fairness conflict.
FPR and FNR describe $\hat y\giv y$,
while PPV describes $y\giv\hat y$. If two
subsets of observations have the same PPV but
different prevalences $p$, then they cannot
also match up on FPR and FNR. Fairness in
$y\giv \hat y$ terms and fairness in $\hat y\giv y$
terms can only coincide in trivial settings
such as when $\hat y=y$ always or empirically unusual
settings with equal prevalence between
observations having different values of a protected
variable.

There is some debate about when or whether
using protected variables can lead to
improved fairness.
See \citet{xian:2020} who gives a summary of
legal issues surrounding fairness
and \citet{corb:etal:2017}
who study whether
imposing calibration or other criteria
might adversely affect the groups they
are meant to help.

\citet{datt:2016} attributes demographic parity of the prediction $\hat y$ 
to input variables through aggregated Shapley value.
We quantify individual level bias on protected variables 
using cohort Shapley and then aggregate to
a measure for any group of interest.
Cohort Shapley can measure quantities of various fairness definitions
that are conventionally quantified by group level statistical measures.

\subsection{COMPAS recidivism data}

The COMPAS Core Risk and Needs Assessment tool from Northpointe Inc.\ includes the General Recidivism Risk Scale (GRRS) and Violent Recidivism Risk Scale (VRRS) which were investigated by \citet{angw:2016}.  A more complete description of these scales can be found in \citet{bren:etal:2009} and \citet{nort:2019}.  We will focus on the GRRS data collected and published by \citet{angw:2016} in the analysis that follows when referring to the ``COMPAS data''.  Each defendant is rated into one of
ten deciles with higher deciles considered higher
risk of reoffending.  \citet{angw:2016} investigated whether that algorithm
was biased against Black people. They obtained data
for defendants in Broward County Florida,
including the COMPAS decile, the defendants' race,
age, gender, number of prior arrests and whether the
crime for which they were charged is a felony
or not. \citet{angw:2016} describe how they
processed their data including how they found
followup data on offences committed and how
they matched those to defendants
for whom they had prior COMPAS scores.
They also note that race was not one of the variables
used in the COMPAS predictions, and the original model for those predictions was trained on more features than those available and included in this Broward County data.  \citet{angw:2016} look at pre-trial defendants and measure recidivism as being arrested for another crime within two years, and we will use the same metrics and data for our analysis so that it can be compared to the original work.  
This example is controversial.  For a more complete discussion of the appropriateness of using this data to evaluate COMPAS including the important discrepancies between pre-trial vs probation or parole use and the time frame for reoffending, see \citet{rudi:etal:2020}, \citet{flor:etal:2016}, \citet{diet:etal:2016}, and \citet{jack:mend:2020}.  While this data is therefore likely inadequate for the purposes of evaluating the true merits of COMPAS, we include it as an example here given its place in the literature as a known benchmark for meaningful comparison.

\citet{angw:2016} find that COMPAS
is biased because it gave a
higher rate of false positives for Black
defendants and a higher rate of false negatives
for White defendants.
\citet{flor:etal:2016} and \citet{diet:etal:2016} disagree, raising the issue of $\hat y\giv y$
fairness versus $y\giv \hat y$ fairness.
The prevalence of reoffences differed
between Black and White defendants
forcing $y\giv\hat y$ and $\hat y \giv y$
notions to be incompatible. \citet{tan:etal:2018} find some evidence of racial bias when comparing an interpretable surrogate model trained to predict $\hat{y}$ to one trained on $y$, but stop short of attributing that bias to COMPAS itself instead pointing to the difference between features used to train COMPAS and those present in the data set. \citet{agar:etal:2021} also find evidence of racial bias in an interpretable surrogate model trained to predict COMPAS scores.  \citet{rudi:etal:2020} find no evidence that the COMPAS scores depend on race except through age and criminal history.  Further they find that the same racial FPR/FNR discrepancies detected by \citet{angw:2016} when looking at the COMPAS scores would exist when using a model that relied solely on age.

Our cohort Shapley analysis below finds statistically significant racial effects.  White defendants tend to have higher than predicted rates of recidivism while Black defendants tend to have lower than predicted rates of recidivism. The magnitudes of those effects are in the range of 3.5 to 5.5 percent when measured by the average of $y-\hat y$. Consistent with some of the earlier findings we see larger effects of race on the false positive and false negative rates.

Beyond these aggregate findings, our analysis provides local (i.e., individual level) racial effects. 
This would be useful in looking at a specific
case of interest or, as we show below, for finding
that the impact of race varies by gender.
While some previous work like \citet{angw:2016} and \citet{rudi:etal:2020} have examined specific individuals or pairs of individuals as examples, this work is the first to present a method that quantifies this local fairness systematically.

Following \citet{chou:2017} we focus on just Black
and White defendants. That provides a sample of
5278 defendants from among the original 6172 defendants.
As in that paper we record the number of prior
arrests 
as a categorical variable with
five levels: 0, 1--3, 4--6, 7--10 and $>$10.
Following \citet{angw:2016}, we record the
defendants' ages as a categorical variable
with three levels:
$<$25, 25--45 and $>$45.
Also, following \citet{chou:2017}, we consider the
prediction $\hat y_i$ to be $1$ if defendant
$i$ is in deciles 5--10 and $\hat y_i=0$
for defendant $i$ in deciles 1--4.

\begin{figure}[t]
  \centering
  \includegraphics[width=1.0 \textwidth]{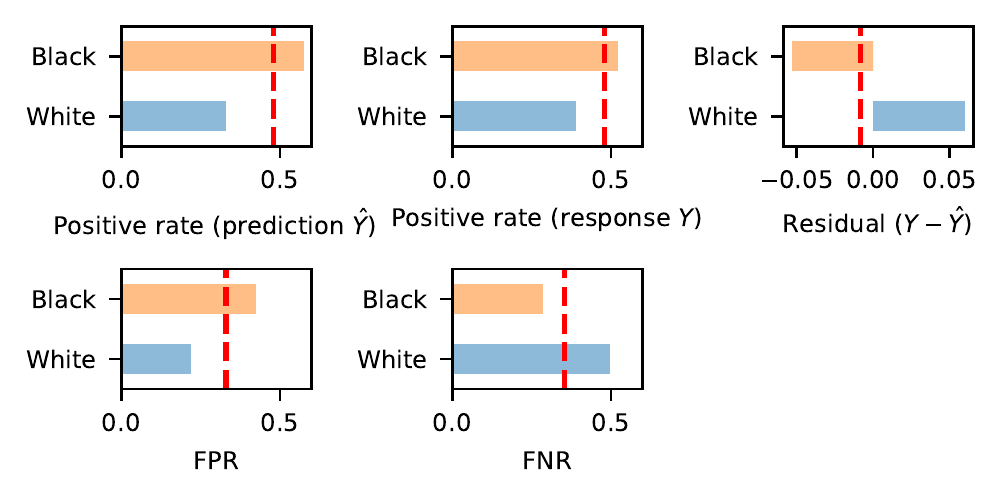}
  \caption{Group fairness metrics on COMPAS recidivism data.}
  \label{fig:compas_gf}
\end{figure}

Figure~\ref{fig:compas_gf} shows some
conventional group
fairness metrics for the COMPAS data set.
Horizontal bars show group specific means
and the vertical dashed lines show population
means.
We see that Black defendants had a higher
average value of $\hat y$ than White defendants.
Black defendants also had a higher average of $y$ but
a lower average residual $y-\hat y$.
Using $B$ and $W$ to denote the two
racial groups,
$\hat\e(y-\hat y\giv B)\doteq -0.035$
and $\hat\e(y-\hat y\giv W)\doteq 0.054$.
The FPR was higher for Black defendants and the FNR
was higher for White defendants.

\subsection{Exploration of the COMPAS data}\label{sec:dataanalysis}

In this section we use cohort Shapley
to study some fairness issues in the COMPAS
data, especially individual level metrics.
We have selected what we found to be
the strongest and most interesting findings
related to race and gender from \citet{fairnessshapley}.
Histograms there also show effects due to age and
the number of prior arrests.

We have computed Shapley impacts for these
responses: $y_i$, $\hat y_i$, $y_i-\hat y_i$,
$\fp_i = 1\{ y_i=0\,\&\, \hat y_i=1\}$ and
$\fn_i = 1\{ y_1=1\,\&\, \hat y_i=1\}$.
If $\fp_i=1$ then defendant $i$ received a
false positive prediction. Note that the
sample average value of $\fp_i$
$$\hat\e(\fp_i)=\frac{n_{01}}{n_{\sumdot\sumdot}}
=\fpr\times \frac{n_{0\sumdot}}{n_{\sumdot\sumdot}}
=\fpr\times (1-p), \text{ and } \hat\e(\fn_i)=\fnr\times p$$

\subsection{Graphical analysis}
\begin{figure}[t]
  \centering
  \includegraphics[width=1.0 \textwidth]{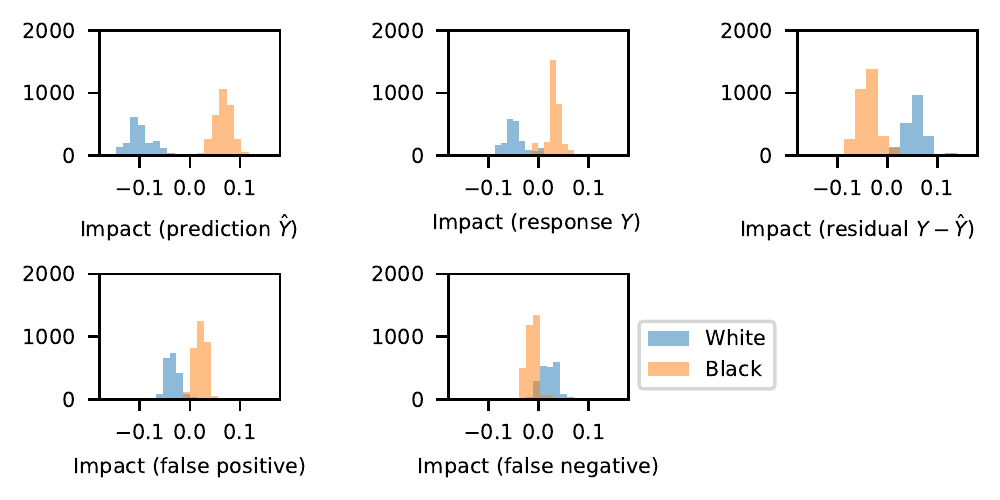}
  \caption{Histogram of cohort Shapley value of race in the COMPAS recidivism data.}
  \label{fig:compas_cs_race}
\end{figure}

Figure~\ref{fig:compas_cs_race} shows histograms
of Shapley impacts of race for the defendants
in the COMPAS data. The first panel there 
shows a positive impact for every Black defendant
and a negative one for every White defendant
for the prediction $\hat y$.
For the actual response $y$, the histograms
overlap slightly. By additivity of Shapley
value the impacts for $y-\hat y$ can be
found by subtracting the impact for $\hat y$
from that for $y$ for each defendant $i=1,\dots,n$.
The histograms of $y_i-\hat y_i$ show that
the impact of race on the residual is typically
positive for White defendants and negative
for Black defendants.

The histograms of Shapley impacts for race overlap for the two metrics, FPR and FNR. There, a small number of adversely affected White defendants and beneficially
affected Black defendants are observed.
We can inspect the corresponding defendants to see which conditions are the exceptional cases. For example, the 469'th defendant (Black) and the 486'th defendant (White) have the same values for all other features. They both had 4-6 prior arrests,
age  $<$25, crime severity of felony, and both were female.
In Section~\ref{sec:fprfnr} we give an example with a confidence interval for the Shapley values of an individual defendant. The same could be done for a difference between two defendant's Shapley values.

\begin{figure}[t]
  \centering
  \includegraphics[width=1.0 \textwidth]{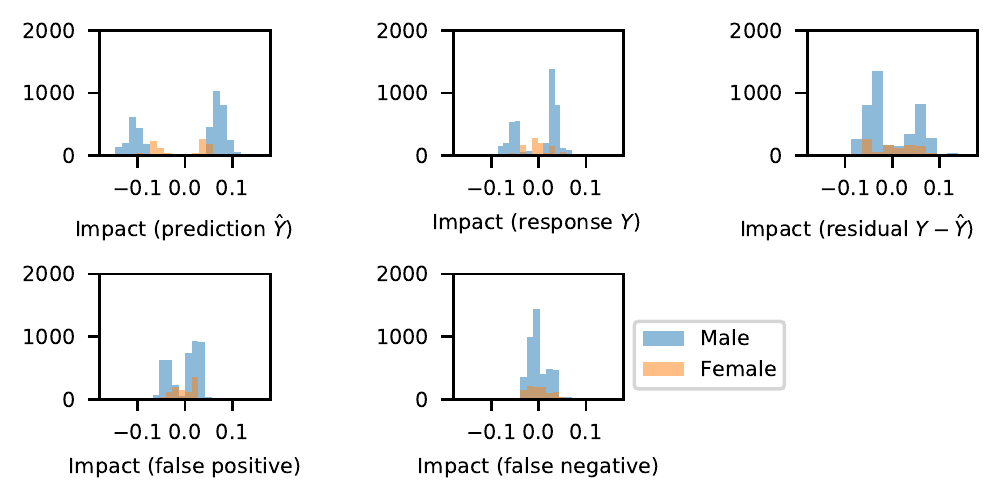}
  \caption{Histogram of cohort Shapley value of race for each gender in the COMPAS recidivism data.}
  \label{fig:compas_cs_race_gender}
\end{figure}

Figure~\ref{fig:compas_cs_race_gender} shows
histograms of Shapley impacts for race color-coded by
the defendants' gender.
We see that the impact on $\hat y$ is bimodal
by race for both male and female defendants,
but the effect
is larger in absolute value for male defendants.
The impacts for the response, the residual
and false positive values do not
appear to be bimodal by race for female defendants,
but they do for male defendants. The impacts for
false negatives do not appear bimodal for either race. 
It is clear from these figures that the
race differences we see are much stronger
among male defendants.

\subsection{Tabular summary and bootstrap}

\begin{table}[t]
\centering
\begin{tabular}{lcccc}
\toprule
Variable & White & Black & Male & Female \\
\midrule
race\_factor & $\phm$0.054 & $-$0.036 & $-$0.001 & $\phm$0.003 \\
gender\_factor & $-$0.001 & $\phm$0.001 & $\phm$0.020 & $-$0.082 \\
\bottomrule\\
\end{tabular}
\smallskip
\begin{tabular}{lcccc}
\toprule
Variable & White-Male & White-Female & Black-Male & Black-Female \\
\midrule
race\_factor & $\phm$0.058 & $\phm$0.042 & $-$0.037 & $-$0.031 \\
gender\_factor & $\phm$0.023 & $-$0.082 & $\phm$0.018 & $-$0.083 \\
\bottomrule\\
\end{tabular}
\caption{\label{tab:csi_residual_subset}
Mean cohort Shapley impact of groups on residual $y - \hat y$.
}
\end{table}

We take a particular interest in the
residual, or error, $y_t-\hat y_t$.
It equals $1$ for false negatives and $-1$
for false positives and $0$ when the prediction
was correct.
Table~\ref{tab:csi_residual_subset} shows
race and gender Shapley values for
the residual $y_i-\hat y_i$ aggregated over all four race-gender subsets of defendants.

What we see there is that revealing that
a defendant is Black tells us that, on average,
that defendant's residual $y_t-\hat y_t$
is decreased by $3.6$\%. By this measure, the Black defendants offend less than predicted. Revealing that
the defendant is White increases the residual
by $5.4$\%. Revealing race makes very little
difference to the residual averaged over
male or over female defendants (of both races).
Revealing gender makes a relatively large difference
of $-8.2$\% for female defendants and $+2.0\%$
for male defendants.

\begin{figure}[t]
  \centering
  \includegraphics[width=1.0 \textwidth]{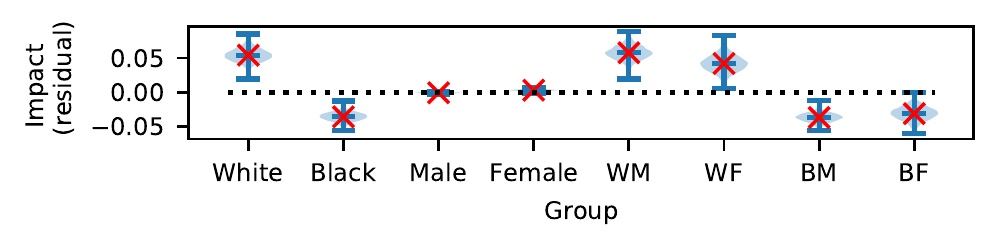}
  \caption{Bayesian bootstrap violin plot of aggregated cohort Shapley race factor impact of groups on residual $y - \hat y$.
  Red crosses show point estimates of mean aggregated cohort
  Shapley values.}
  \label{fig:agg_race_impact_violin}
\end{figure}

To judge the uncertainty in the values in
Table~\ref{tab:csi_residual_subset}
we applied the Bayesian bootstrap of \citet{rubi:1981}.
That algorithm randomly reweights each data point
by a unit mean exponential random variable. There is an asymptotic justification in \citet{newton1994approximate}.
Figure~\ref{fig:agg_race_impact_violin} shows
violin plots of 1000 bootstrapped cohort Shapley values. 
\newt{We see that the mean Shapley value of race on
the residual is significantly negative for Black defendants and significantly positive for White defendants. These differences persist when disaggregated by gender.  The average effect of race is near zero for male defendants and for female defendants due to race effects nearly cancelling.}

\begin{figure}[t]
  \centering
  \includegraphics[width=1.0 \textwidth]{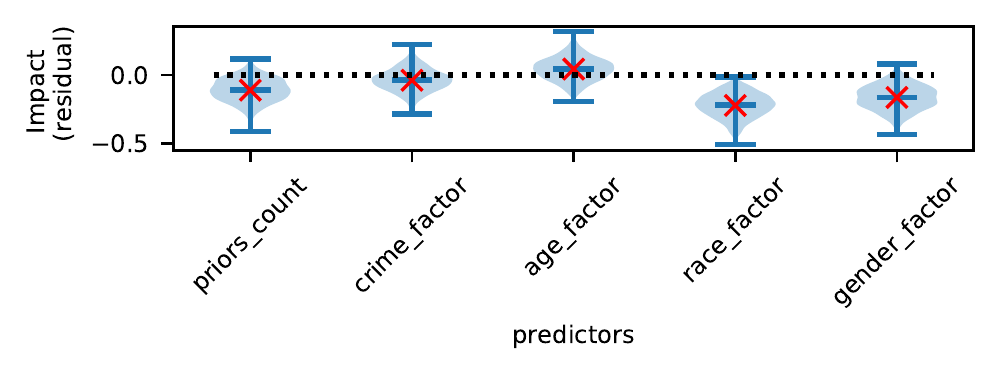}
  \caption{Bayesian bootstrap violin plot of cohort Shapley values for the residual $y-\hat y$ score for defendant $t=2999$.
  Red crosses show the cohort Shapley point estimates for this defendant. }
  \label{fig:idx2999_res_violin}
\end{figure}

Because the Bayesian bootstrap never fully deletes any of the data we can use it to quantify the statistical uncertainty in the cohort Shapley values for an individual defendant.
Figure~\ref{fig:idx2999_res_violin} shows a Bayesian bootstrap violin plot of cohort Shapley values on the residual for the 2999'th defendant. This is the Black defendant whose Shapley value for race on the residual $y-\hat y$ was most negative, in the region overlapping Shapley values of White defendants. As one would expect there is greater uncertainty on the impacts for an individual than for an aggregate. Our analysis has not taken account of the fact that this defendant was selected based on their data. 

\subsection{FPR and FNR revisited}\label{sec:fprfnr}

The risk of being falsely predicted
to reoffend involves two factors:
having $y_i=0$ and having $\hat y_i=1$
given that $y_i=0$. FPR is commonly
computed only over defendants with $y_i=0$.
Accordingly, in this section we study
it by subsetting the defendants to
$\{ i\mid y_i=0\}$ and finding cohort Shapley
value of $\hat y_i=1$ for the features. That provides a different way to quantify the importance of race on the incidence of false positives.

\begin{figure}[t]
  \centering
  \includegraphics[width=1.0 \textwidth]{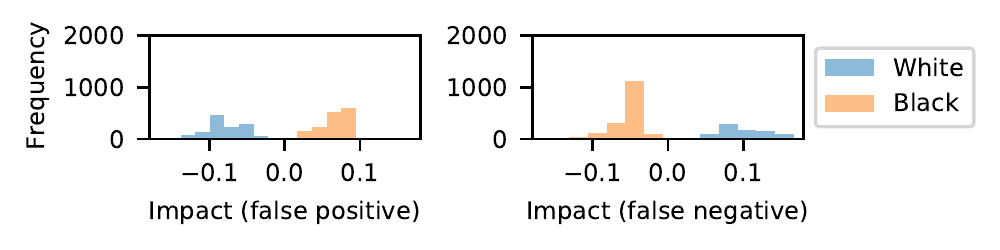}
  \caption{Histogram of cohort Shapley values of race  in the COMPAS recidivism data. For false positives, only defendants with $y_i = 0$ are included. For false negatives, only defendants with $y_i = 1$ are included.}
  \label{fig:compas_cs_fprfnr}
\end{figure}

Figure~\ref{fig:compas_cs_fprfnr} shows the cohort Shapley impact of race conditioned for the FPR and FNR
after working with subsets of defendants as described above.
The distribution of impacts conditioned on race are clearly separated in the figure.
This conditional analysis shows a stronger disadvantage for Black defendants than in Figure~\ref{fig:compas_cs_race}.

\section{Discussion}\label{sec:discussion}

There are many incompatible definitions of importance. Local importance measures address a causes of effects question, which means that we have to carefully consider which counterfactuals to use and how to use them.
When attributing the value of $f(\bsx_i)$ to features $x_{ij}$, we have avoided using any impossible data by only using actually observed data. We think it is important to have such a choice as otherwise the attribution could be based on extremely unlikely or even impossible combinations. \citet{hook:ment:2019:tr} also point to problems with mixing and matching variable inputs.

 There could be good uses for possible but hitherto unobserved input combinations. Using only observed values means that we rely on data cleaning to have eliminated any impossible combinations. An alternative that uses input values that pass a `within distribution' test could be used but in our view it is exceedingly difficult to draw the line between possible and impossible values based on a sample. \citet{kuma:2020} also raise this point.

There may be a use for impossible data when explaining whether a predictor affects accuracy. For instance, when judging whether a predictor helps predictions, \citet{brei:2001} studies what happens when $x_{1j},\dots,x_{nj}$ are subjected to a random permutation $\pi(\cdot)$ with respect to the other variables, producing hypothetical values $\tilde\bsx_i=\bsx_{i,-j}{:}\bsx_{\pi(i),j}$. He then looks at the accuracy of predicting $y_i$ by $f(\tilde\bsx_i)$. If accuracy does not change much, then it is reasonable to suppose that $x_j$ has only a minimal role in predictions. If accuracy does change a lot however, it could be because the fitted model extrapolates poorly into the low probability or impossible regions of the space. Breiman's method appears then to be a `one way diagnostic', persuasive when it indicates low importance but otherwise not.

Despite these possibilities, we adopted a constraint of not using any unobserved values in order to define a measure of importance.  As mentioned in the introduction this counters adversarial approaches that can make changes to predictions at unobserved points in order to change the fairness measures of an algorithm.

In defining cohort Shapley we have required a human in the loop to define a similarity measure for the variables.  We acknowledge that this is a burden. However, the alternatives we have seen use additional black boxes or use the response itself to define similarity and we think those choices make the final result harder to understand.  We think that people
would not accept an automatic algorithmic
declaration that some black box is or is not fair
without having the ability to see how it defines
the similar cases that should be treated similarly.
Also, the human or humans making and discussing those choices can decide on similarity for one variable at a time.  Those choices then define the conditional expectations needed by the algorithm.

In addition to transparency and avoiding impossible data, cohort Shapley brings some other advantages: the user can opt to measure importance of variables that were not used in the model, the method does not require access to the black box beyond the predictions it makes, and having defined cohorts we can study means or medians or other quantities of interest. \newt{It can also be applied to data on which the algorithm was not trained, as for instance with the COMPAS data. This allows one to quantify external validity properties directly.}   All conditional expectation Shapley value methods use estimates of conditional means and these become challenging as the conditioning dimension increases and the sample size decreases. For a setting where the black box is used on fresh data that it was not trained on, we have illustrated a convenient bootstrap method to quantify the uncertainty that arises from sampling observations.

With the COMPAS data we have illustrated how cohort Shapley values can be used to explore data surrounding a complex fairness issue. There are many more analyses that could be done, and we expect that domain experts would not be unanimous on which is best. We do not have the data necessary to compare fairness of COMPAS to fairness of human judges. 

\section*{Acknowledgments}
This work was supported by Hitachi, Ltd. and by the National Science Foundation grant IIS-1837931. We thank Sharad Goel, Jessica Hwang, Alexandra Chouldechova and Alice Xiang for helpful discussions. The opinions in this article are our own and not necessarily shared by those we acknowledge.

\bibliographystyle{apalike}
\bibliography{references}

\end{document}